\documentclass{vldb}
\usepackage{graphicx}
\usepackage{balance}  

\usepackage{algorithm}
\usepackage[noend]{algpseudocode}

\usepackage[utf8]{inputenc}
\usepackage{adjustbox}
\usepackage{multirow}
\usepackage{verbatim,multirow,mathtools,enumitem,textcomp,capt-of,url,color,lipsum,epstopdf,graphicx,relsize,balance,times,layouts,amssymb}

\newtheorem{definition}{Definition}[section]
\newtheorem{example}{Example}[section]

\usepackage[normalem]{ulem}
\usepackage{amsmath}
\DeclareMathOperator*{\argmax}{arg\,max}

\newcommand{\eat}[1]{}

\newcommand{\ie}{{\em i.e.}}
\newcommand{\eg}{{\em e.g.}}

\newcommand{\tightlist}{\itemsep=0.5pt}
\newcommand{\page}{$w$ }

\newcommand{\demcaps}{CERES}
\newcommand{\demabbrev}{CERES}

\usepackage{booktabs}
\usepackage{array}
\newcolumntype{L}[1]{>{\raggedright\let\newline\\\arraybackslash\hspace{0pt}}m{#1}}
\newcolumntype{C}[1]{>{\centering\let\newline\\\arraybackslash\hspace{0pt}}m{#1}}
\newcolumntype{R}[1]{>{\raggedleft\let\newline\\\arraybackslash\hspace{0pt}}m{#1}}
\usepackage{float}

\usepackage[T1]{fontenc}
\usepackage[utf8]{inputenc}

\begin{document}


\title{CERES: Distantly Supervised Relation Extraction \\from the Semi-Structured Web}



%
%
%
%

\numberofauthors{4} 

\author{
%
%
\alignauthor
Colin Lockard\titlenote{\label{amazon}All work performed while at Amazon.}\\
       \affaddr{University of Washington}\\       
       \email{\small lockardc@cs.washington.edu}
\alignauthor
Xin Luna Dong\\
       \affaddr{Amazon}\\
       \email{\small lunadong@amazon.com}       
\\\and
\alignauthor 
Arash Einolghozati\titlenote{}\\
       \affaddr{Facebook}\\
       \email{\small arashe@fb.com}       
\alignauthor
Prashant Shiralkar\\
      \affaddr{Amazon}\\
      \email{\small shiralp@amazon.com}
}
\date{30 July 1999}

\maketitle

\begin{abstract}
The web contains countless semi-structured websites, which can be a rich source of information for populating knowledge bases. Existing methods for extracting relations from the DOM trees of semi-structured webpages can achieve high precision and recall only when manual annotations for each website are available. Although there have been efforts to learn extractors from automatically-generated labels, these methods are not sufficiently robust to succeed in settings with complex schemas and information-rich websites. 

In this paper we present a new method for automatic extraction from semi-structured websites based on distant supervision. We automatically generate training labels by aligning an existing knowledge base with a web page and leveraging the unique structural characteristics of semi-structured websites. We then train a classifier based on the potentially noisy and incomplete labels to predict new relation instances. Our method can compete with annotation-based techniques in the literature in terms of extraction quality. A large-scale experiment on over 400,000 pages from dozens of multi-lingual long-tail websites harvested 1.25 million facts at a precision of 90\%.

\end{abstract}

\section{Introduction}
\label{sec:introduction}
Knowledge bases, consisting a large set of (subject, predicate, object) {\em triples} to provide factual information, have recently been successfully applied to improve many applications including Search, Question Answering, and Personal Assistant. It is critical to continuously grow knowledge bases to cover long tail information from different verticals (\ie, domains) and different languages~\cite{Li2017}, and as such, there has been a lot of work on automatic knowledge extraction from the Web~\cite{Dong2014KnowledgeVA, Dalvi2011AutomaticWF, Gentile2015EarlyST, FGG+2014, CBK+2010}. 

Among various types of Web sources, we argue that semi-structured websites (\eg, IMDb, as shown in Figure~\ref{fig:DoTheRightThing}) are one of the most promising knowledge sources. The Knowledge Vault project~\cite{Dong2014KnowledgeVA} reported that after applying automatic knowledge extraction on DOM trees of semi-structured websites, texts, web-tables, and semantic web annotations, 75\% of the extracted facts and 94\% of the high-confidence facts were covered by DOM trees. This is not surprising: semi-structured websites are typically populated by data from large underlying databases, thus containing richer information than webtables and web annotations; the underlying databases often focus on factual information, making it more suitable as knowledge sources than free texts.

Despite the huge opportunities, automatic knowledge extraction from semi-structured sources
has not received the attention it deserves from the research community. 
Unlike natural language text, semi-structured sources provide (arguably) more structure, but in a different way, with fewer well-developed tools for tasks like named-entity recognition and entity linking. Unlike webtables~\cite{CHW+2008}, semi-structured data lack the table structure (rows and columns) that helps identify entities and relations. Semi-structured sources also present challenges for automatic extraction because the structure, or layout, differs from website to website, so the extraction model trained for one website cannot be used for another. Even between webpages that are generated from the same template, pages may differ due to missing fields, varying numbers of instances, conditional formatting, and ad and recommendation strategies. 

Traditional DOM extraction typically uses {\em wrapper induction}: given a website, wrapper induction asks for manual annotations, often on only a handful of  pages, and derives the extraction patterns, usually presented as XPaths, that can be applied to the whole website~\cite{Kushmerick1997WrapperIF}. Although wrapper induction has been quite mature in terms of extraction quality, obtaining precision over $95\%$~\cite{Gulhane2011WebscaleIE}, it requires annotations on every website, an expensive and time-consuming step if one wishes to extract from many websites.

In order to create an automated process requiring no human annotation, distant supervision has been proposed for text extraction~\cite{Mintz2009DistantSF}. In this process, training data are generated automatically by aligning sentences with existing seed knowledge bases using simple heuristics; that is, if a sentence mentions two entities of a triple in the knowledge base, the sentence is annotated to assert the relation in the triple. Although the training data can be noisy, the large number of annotations on a large corpus of texts still enables learning fairly robust extraction models. 

Unfortunately, distant supervision does not easily apply to semi-structured websites. This is mainly because the unit for annotation changes from a sentence to a webpage, normally containing much more information. The sheer volume of text on a webpage first poses efficiency challenges: a single web page may contain mentions of hundreds of entities, which can match to thousands of potential candidate entities; examining relations between every pair of candidate entities can be very expensive. Even worse, the large number of entities mentioned in a webpage may cause many spurious annotations, leading to low-quality extraction models. Indeed, a recent attempt to apply distantly supervised extraction on semi-structured data obtained quite low accuracy: Knowledge Vault trained two distantly supervised DOM extractors but their accuracy is only around 63\%~\cite{Dong2014FromDF}.

In this paper, we present \demcaps\footnote{Ceres is the Roman goddess of the harvest.}, a knowledge extraction framework that improves the distant supervision assumption for DOM extraction. The key underlying idea is to best leverage the unique characteristics of semi-structured data to create fairly accurate annotations. First, we focus on detail pages, each of which describes an individual entity, and propose an automatic annotation algorithm that proceeds in two steps: it first identifies the \textit{topic entity}, which is the primary subject of a page, and then annotates entities on the page that are known (via the seed KB) to have relationships with that entity. This reduces the annotation complexity from quadratic to linear and drops a significant number of spurious annotations. Second, we leverage the common structure between key-value pairs within a webpage, and across webpages in a website, to further improve annotation quality. 

Our new extraction method significantly improves extraction quality. On the SWDE dataset~\cite{Hao2011FromOT}, which has been used as a standard testbed for DOM extraction, we are able to obtain an average accuracy of over 90\% in various verticals, even higher than many annotation-based wrapper induction methods in the literature. Large-scale experiments on over 400,000 pages from dozens of multi-lingual long-tail websites harvested 1.25 million facts at a precision of 90\%, with the ratio of 1:2.6 between annotated popular entities and extracted entities that include many long-tail entities.

Our paper makes the following contributions.
\begin{enumerate}
  \item We describe \demcaps, an end-to-end distantly supervised knowledge extraction framework that can apply to semi-structured websites independent of domains and web sources.\footnote{We note that this approach can be combined with a manual-annotation-based approach; when entering a new domain for which no KB exists, an annotation-based extractor could be run on a few prominent sites and used to populate a seed KB for distantly supervised extraction of other sites.}
  \item We propose an advanced distant supervision annotation process for semi-structured web sources by leveraging the unique structural characteristics of such data. 
  \item In addition to showing the efficacy of our technique on a benchmark dataset, we include an evaluation on dozens of real-world long-tail websites containing multi-valued predicates to show applicability of our method in the wild.
\end{enumerate}

The rest of the paper is structured as follows. Section~\ref{sec:overview} formally defines the problem and overviews our solution. Section~\ref{sec:annotation} describes data annotation and Section~\ref{sec:learning} describes training and extraction. Section~\ref{sec:experiment} presents our experimental results. Section~\ref{sec:related} reviews related work and Section~\ref{sec:conclude} concludes.
 
\section{Problem Definition and Solution Overview}
\label{sec:overview}
We introduce the problem in this section and illustrate the challenges (Section~\ref{subsec:defn}). We then review distant supervision (Section~\ref{subsec:distant_supervision}), and give an overview of our solution (Section~\ref{subsec:overview}).

\subsection{Problem definition}
\label{subsec:defn}

\noindent
\textbf{KB:} We assume a setting in which we have an existing seed \emph{knowledge base} (KB) populated with at least some facts corresponding to a given ontology. Facts are represented by \emph{triples} of the form $(s, r, o)$, where $s$ is the {\em subject} of the relationship, $r$ is the relation {\em predicate} according to our ontology, and $o$ is the {\em object} of the relationship. For example, a triple expressing the relationship between the film \emph{Do the Right Thing} and its director {\em Spike Lee} would be \textit{(Do the Right Thing, directed by, Spike Lee)}. Some predicates participate in a single triple with a subject, specifying a unique value (such as a birth date), while others are \textit{multi-valued} and may participate in many triples with one subject, such as the predicate indicating an actor has acted in a film. The \emph{ontology} defines the semantics of the relation predicates. Our goal is to further populate the KB with facts obtained from semi-structured websites. 

\begin{figure}[t!]
\includegraphics[width=8cm]{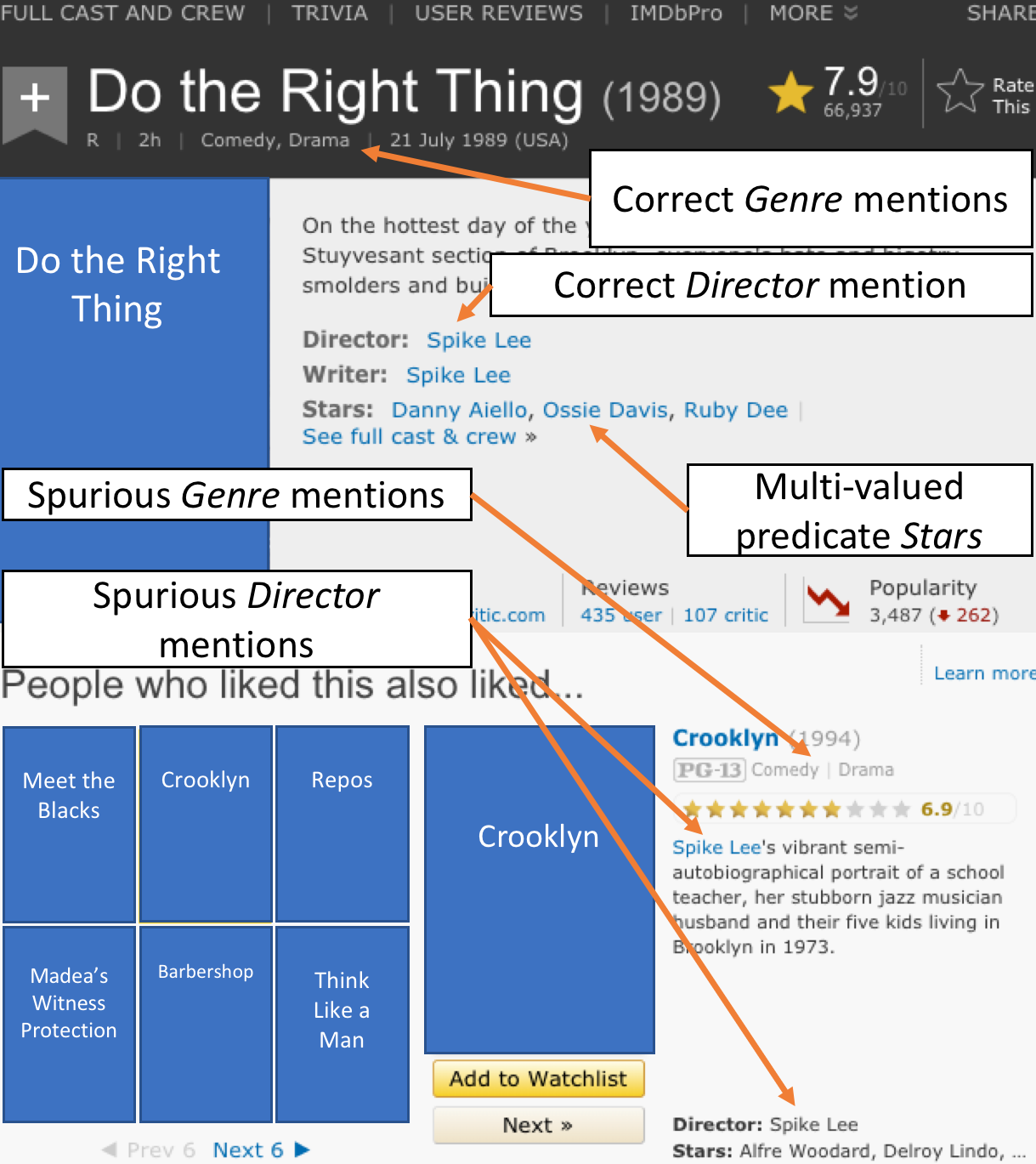}
\caption{A cropped portion of the detail page from imdb.com for the film \emph{Do the Right Thing}. Labels indicate some of the annotation challenges.}
\label{fig:DoTheRightThing}
\vspace{-1.2em}
\end{figure}

\smallskip
\noindent
\textbf{Semi-structured website:} A \emph{semi-structured website} $W$ consists of a set of \emph{detail pages} that have been generated from a template or a set of templates, with the pages from the same template sharing a common structure and placement of information. Each detail page contains one or more facts about a particular entity, called the {\em topic entity} of the page. An example of a semi-structured website is the Internet Movie Database (IMDb)\footnote{www.imdb.com}, which contains thousands of pages about movies, with each page giving information such as the title, director, writer, release date, and stars of a particular film. An example page is shown in Figure \ref{fig:DoTheRightThing}. 
In this paper we are only interested in extraction from detail pages, not website entry pages or search pages; we consider it likely that every entry in an entry page corresponds to some detail page that provides more information.

\begin{figure}[t]
\includegraphics[width=8cm]{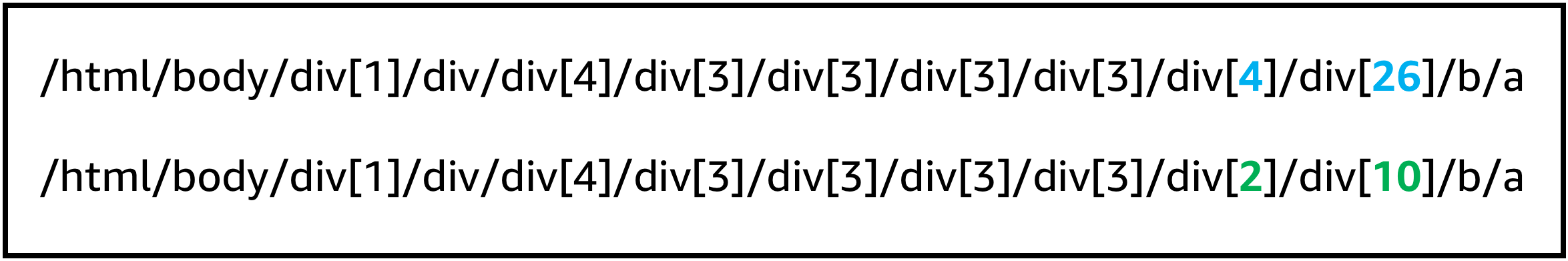}
\caption{Example of XPaths corresponding to the \textit{acted in} predicate on two IMDb pages. They differ at two node indices, and the second path corresponds to the \textit{producer of} predicate from the first page.}
\label{fig:tree_path}
\vspace{-1.2em}
\end{figure}

While each page on a semi-structured website is expected to share a common structure, there may be minor differences from page to page. Some pages may be missing some fields, either because that field does not exist (such as date of death for a person who is still living) or because the database underlying the website lacks the information. 
There may also be variations in structure because of variations in the number of instances of a predicate (for example, most films on IMDb have only a single director, but some films have multiple directors). 
In order to extract from a semi-structured website, an extractor must be robust to these changes in the placement of facts.

We represent each webpage as a \textbf{DOM Tree}\footnote{See https://www.w3.org/DOM/}; a node in the tree can be uniquely defined by an absolute XPath, which for brevity we simply refer to as an \textbf{XPath} \cite{Olteanu2002XPathLF}. XPaths representing the same predicate on a semi-structured website tend to be similar but may have some differences. Figure \ref{fig:tree_path} shows an example of XPaths representing the ``Acted In'' predicate on IMDb. The upper path is from Oprah Winfrey's page whereas the lower path is from Ian McKellon's. On Winfrey's page, a ``Producer'' section exists in the section corresponding to the Actor section on McKellon's page, pushing her film performances down, altering the second-to-last index. Both performers act in many films, with the final index indicating the placement of the node in the film list. Notably, the upper XPath from McKellon's page does exist on Winfrey's page as well, but it represents her ``producer of'' relationship with the film \textit{Selma}.

\smallskip
\noindent
\textbf{Problem Definition:} Our goal is to identify the DOM nodes of a detail page containing facts corresponding to particular predicates in our ontology. 

\begin{definition}[DOM Extraction]
Let $W$ be a semi-structured website and $K$ be a seed knowledge base on a particular vertical. {\em DOM extraction} extracts a set of triples from $W$ such that the subject and object of each triple is a string value on a page in $W$, and the predicate appears in $K$ and indicates the relation between the subject and object as asserted by the webpage.

\end{definition}

A successful extraction shall extract {\em only} relations that are asserted by the website (high precision) and extract {\em all} relations that are asserted (high recall). We consider only predicates in the ontology, for which we can obtain training data from $K$. 
Note that we resort to existing work to verify the accuracy of the website's claims~\cite{DGM+2015} and to tackle the problem of entity linkage between the extracted data and existing entities in the knowledge base~\cite{BDIBook}.  We observe that most entity names correspond to full texts in a DOM tree node, so for simplicity we leave identifying entity names as substrings of texts in a DOM node or concatenation of texts from multiple DOM nodes to future work. 

Finally, we clarify that even in the same website, semi-structured webpages may be generated by very different templates; for example, on IMDb, movie pages and people pages appear quite different. We first apply the clustering algorithm in~\cite{Gulhane2011WebscaleIE} to cluster the webpages such that each cluster roughly corresponds to a template, and then apply our extraction algorithm to each cluster. 

\subsection{Review of distant supervision}
\label{subsec:distant_supervision}
The key challenge in extracting knowledge from 
the semi-structured web
without human intervention
is to automatically generate training data
for each website;
{\em distant supervision} has been proposed for this task. Distantly supervised relation extraction is a process that first aligns a knowledge base to an unannotated dataset to create annotations, and then uses the annotations to learn a supervised extraction model. Distant supervision was initially proposed for relation extraction from natural language text and in that context relies on the \textit{distant supervision assumption}: if two entities are known (via the seed KB) to be involved in a relation, any sentence containing both entities expresses that relation \cite{Mintz2009DistantSF}. Because this is obviously not true in all cases, much of the work in distantly supervised extraction has been devoted to learning effectively despite the noisy labels generated by this assumption. 

The distant supervision assumption is harder to apply in the DOM setting for three reasons. First, unlike natural language text, where the basic unit of analysis is a sentence that usually consists of no more than a few dozen words and mentions only a few entities, in DOM extraction the basic unit of analysis is a web page. Web pages may contain thousands of text fields, and many real-world use cases will mention hundreds or thousands of entities, which poses challenges both at annotation time and at extraction time. 

To understand the difficulties in successfully annotating relation mentions in this setting, consider the IMDb page of prolific voice actor Frank Welker\footnote{\small \url{http://www.imdb.com/name/nm0919798/}}. Welker's credits list over 800 films and TV series in which he has performed. The page also lists Welker's place and date of birth, characters he has portrayed, television episode titles, and various other entities and attributes, which may exist in our KB. This single page contains legitimate mentions of several thousand entities which may have relationships with Frank Welker, with some of them mentioned multiple times. 

Second, the situation is further complicated by the fact that strings appearing on a web page may correspond to multiple entities in our KB, sometimes many entities. For example, the word ``Pilot'' is the name of thousands of television episodes, since it is used for the first episode of a series. In addition, other strings may incidentally match entities in our KB; for example, Welker's page has a section labeled ``Biography'', which also happens to be the title of a TV series. Other sections of the page may reference entities unrelated to the topic entity; for example, the detail page about the film \emph{Do the Right Thing} shown in Figure \ref{fig:DoTheRightThing} also mentions information about \emph{Crooklyn}, including several of its cast members. 

The result of this setting is that a single web page may contain strings corresponding to tens of thousands of entities in our KB, and hundreds or thousands of those matches may even be correct. The distant supervision assumption would have us check our KB for relationships involving all pairs of entity mentions, and if a relation was found, would have us produce an annotation for that predicate for those fields of the page. This approach is both computationally challenging, due to the immense number of potential entity pairs, as well as likely to produce noisy annotations due to the spurious matches made likely by the huge number of potential entities.

Finally, because the distant supervision assumption does not incorporate the knowledge that a detail page contains many relationships involving the same subject (the topic entity of the page), it involves labeling pairs of nodes as expressing a triple. At extraction-time, it will then be necessary to consider all pairs of nodes containing potential entity mentions as targets for extraction. However, unlike in NLP,  pre-trained NER systems are not available in the DOM setting, and thus it is not obvious how to select these nodes. Considering all possible pairs of nodes on a page is computationally infeasible, while relying on the KB for entity matching, as we do at annotation-time, would limit the ability to produce extractions involving previously unknown entities.

\subsection{Overview of our approach}
\label{subsec:overview}

\begin{figure}[t]
\centering
\includegraphics[width=\linewidth,scale=0.7]{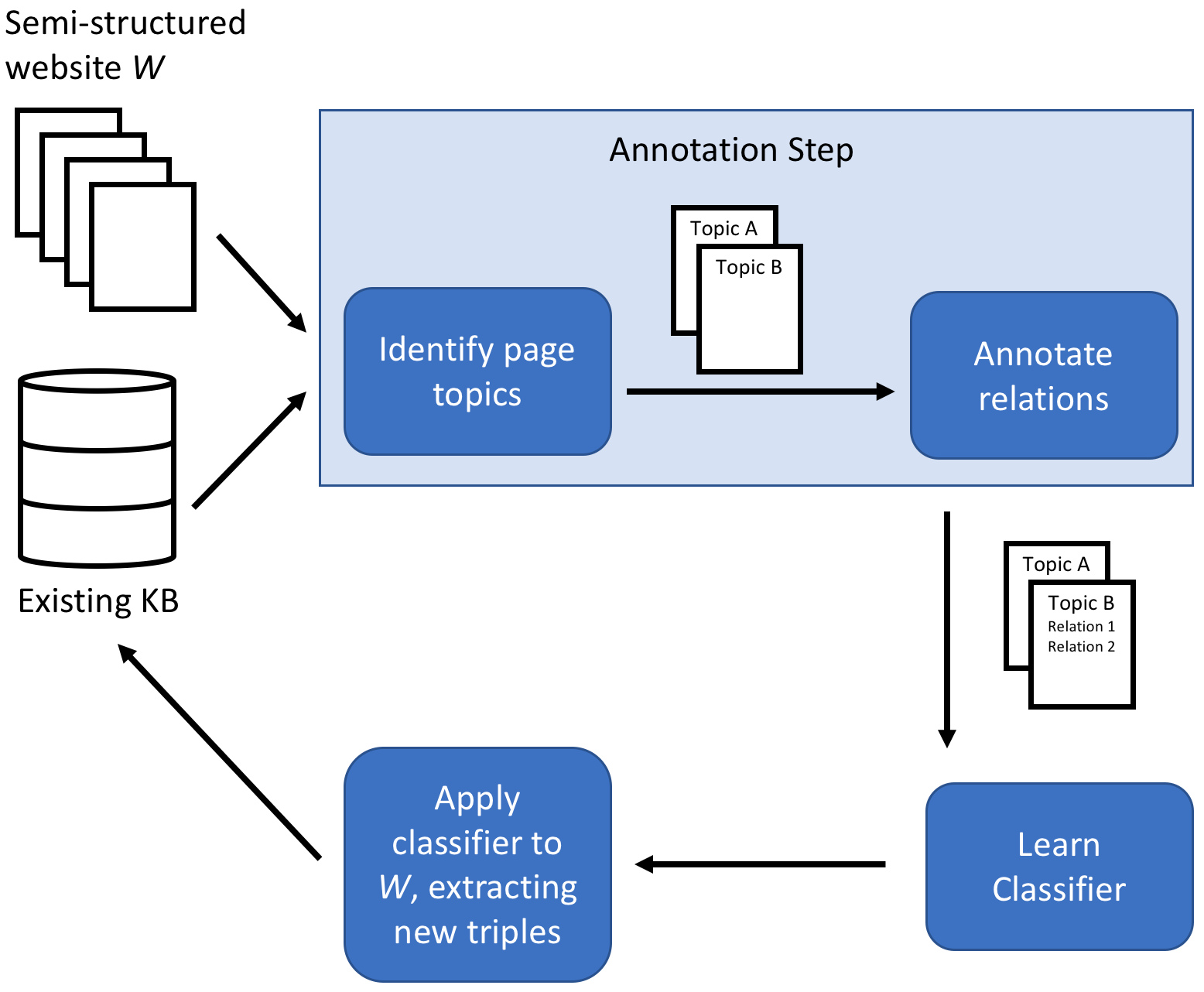}
\caption{{\demcaps }  architecture.}
\label{fig:ProcessDiagram}
\vspace{-1.2em}
\end{figure}

We propose \demcaps, a new method for automatically extracting information from semi-structured webpages. {\demcaps } makes use of an existing knowledge base to automatically label facts on the webpages that can be mapped to known facts in our KB. These annotations are then used to train a probabilistic extractor which extracts additional facts from the webpages. Our process thus involves an annotation step, a training step, and an extraction step, as shown in Figure \ref{fig:ProcessDiagram}. There are three key differences between our approach and a traditional approach to distant supervision.

First, we rely on a modification of the distant supervision assumption which we term the \textit{Detail Page Distant Supervision Assumption}.

\begin{definition}[Detail Page DS Assumption]
\vspace{-0.5em}
Let $w$ be a detail page, and $(s, r, o) \in K$ be a knowledge triple. If $s$ is the topic entity for $w$ and $w$ contains object $o$, it is assumed that \textit{at least one} mention of $o$ on $w$ expresses the relation $r$. 
\vspace{-0.5em}
\end{definition}

According to this assumption, we use a two-step annotation process in which we first attempt to identify the topic entity of a page, and then annotate relations.

Second, text allow great flexibility in expression, with sentences structured differently from document to document. In contrast, webpages in a semi-structured website are expected to share a common structure. As we annotate, we consider both {\em local} evidence within a page, and {\em global} evidence across pages to improve annotation quality.

Third, traditional distantly supervised extraction relies on NER techniques to identify entities, and uses syntax and lexical features to predict relations between every pair of entities in the same sentence. As we extract knowledge from a detail page, we consider only relations between the topic entity and other entities on the page. We use the (possibly noisy) annotations to train a machine learning model that classifies a DOM node, represented by an XPath, on a webpage, and outputs a relation between the topic entity and the entity represented by the node (either a predicate present in the training data or an ``OTHER'' label indicating that there is not a relation present in our ontology). 

We admit that in this way we will not be able to extract relations between non-topic-entities, such as the {\em directed by} relation between the {\em Crooklyn} movie and its director in Figure~\ref{fig:DoTheRightThing}. However, we believe such relations, if important, are likely to be covered by another detail page with the subject as the topic entity (\ie, the webpage for the {\em Crooklyn} movie).

\begin{algorithm}[t]
\caption{Page Topic Identification}
\label{alg:topic}
\begin{algorithmic}[1]
\small
\Require{A set of $n$ semi-structured webpages $W$, and a knowledge base consisting of a set of triples, $K$.}
\Ensure{$W'$ and $T$, a list of pages in $W$ in which a topic was identified, and the corresponding topic entities. }
\Procedure{PageTopicIdentification}{W, K}
\Procedure{ScoreEntitiesForPage}{w}
	\State $p \gets \{\} $ {/* initialize dictionary of entity score */}
	\State $\textit{pageSet} \gets$ all entities on \page via string matching
\ForAll{$e \in PageSet$}
\State {$\textit{entitySet} \gets$ all objects of triples in $K$} 
\State {\hspace{8ex}where $e$ is the subject}
\State $p[e] \gets J(entitySet, PageSet)$
\EndFor
\State \textbf{return} $p$
\EndProcedure
\\
\State {/* Score entities on a page and track counts of XPaths leading to highest scoring entities. */}
\State $P \gets \{\}$ {/* track entity scores for each page */}
\State $pathCounts \gets \{\}$ {/* track XPath counts across $W$ */}
\ForAll{$w_i \in W$}
\State $P[w_i] \gets ScoreEntitiesForPage(w_i)$
\State $c_i \gets \argmax(P[w_i])$
\State {$candidateTreePaths \gets$ XPaths of all mentions }
\State {\hspace{8ex}of $c_i$ on $w_i$}
\ForAll{$path \in candidateTreePaths$}
\State $pathCounts[path] ++$
\EndFor
\EndFor
\\
\State {/* Match text at the most common XPath across $W$ for the highest scoring entity $c_i$ for each page */}
\State $T \gets \{\}$ {initialize dictionary of topic entities}
\ForAll{$w_i \in W$}
\State {$maxCountTreePath \gets $ XPath extant on $w_i$ with}
\State {\hspace{8ex}max count in $pathCounts$}
\State {$topicTextField \gets$ text field found at}
\State {\hspace{8ex}$maxCountTreePath$ on $w_i$}
\State {$T[w_i] \gets K$'s entity string matching $topicTextField$}
\State {\hspace{8ex}with max score in $P(w_i)$}
\EndFor
\State \textbf{return} $T$
\EndProcedure
\end{algorithmic}
\end{algorithm}

\section{Automatic Annotation}
\label{sec:annotation}
The purpose of the annotation process is to produce a set of examples that will be used to train a supervised extractor. With that in mind, we place a high value on maintaining the precision of our annotations even if it prevents us from annotating some pages. To maintain high precision, we base our annotation process on three observations.
\begin{enumerate}
\item \textbf{Consistency:} Detail pages on a website often have roughly the same format, with information expressed in a similar way across pages.  
\item \textbf{Informativeness:} Each detail page typically provides multiple attributes of an entity, and the values are typically fairly diverse among entities (\ie, it seldom happens that all entities across the website share the same object for a predicate).
\item \textbf{Uniqueness:} An entity is typically described by a single detail page or at most a couple of detail pages (some with more data in particular aspects). 
\end{enumerate}

These observations will guide us in our annotation process. We re-state the two key differences from annotation in traditional distant supervision. First, we will identify the topic entity for each page, before annotating all objects of predicates shown on the page. Second, in both steps we will harness the semi-structured nature of the data by relying on global information gathered across all pages of a website in addition to the local information on each page.

\subsection{Topic Identification}
\label{sec:topic_identification}
Given a set of $n$ webpages $w_{1}, ..., w_{n}$  from a semi-structured website $W$, our first step is to identify the topic entity of each page when it applies,
as presented in Algorithm \ref{alg:topic}. 

\subsubsection{Local Topic Candidate Identification}
Local candidate identification proceeds in two steps: it first identifies all entities on the page, and then finds the one that is most likely to be the topic entity. Our intuition is that since the purpose of the detail page is to state facts about the topic entity, more strings on the page should bear a relation to the topic entity than to other entities.

\smallskip
\noindent
{\bf Step 1. Entity identification.} 
We begin this process by attempting to identify all entities that may be listed on the page. Each text field on the webpage is matched against the KB using fuzzy string matching process presented in~\cite{Gulhane2010ExploitingCR}. This yields a set of all entities in the KB that may be mentioned on the page. We will refer to this set of $m$ potential entities on $w_i$ as the $pageSet$. As noted previously, this could contain thousands or tens of thousands of entities, each of which is a potential topic entity of the page.

According to our {\em Uniqueness} observation, we compile a list of strings appearing in a large percentage (\eg, 0.01\%) of triples and do not consider them as potential topics. In addition, we discard strings with low information content, such as single digit numbers, years, and names of countries. 

\smallskip
\noindent
{\bf Step 2. Topic candidate identification.}
For each entity $s_j$ in $pageSet$, we then look in the KB to get a set of entities that are the object of a relation in which $s_{j}, j \in [1,m],$ is the subject; we will refer to this set as $entitySet_{j}$.
We then compute the Jaccard similarity between $pageSet$ and $entitySet_{j}$ for each $e_{j} \in pageSet$:
\begin{equation}
J(pageSet,entitySet_{j}) = \frac{|pageSet \cap entitySet_{j}|}{|pageSet \cup entitySet_{j}|}
\end{equation}

The initial topic candidate $c_{i}$ for page $w_{i}$ is then the entity in $pageSet$ with maximum Jaccard similarity:

\begin{equation}
c_{i} = \argmax_j J(pageSet,entitySet_{j})
\end{equation}

Note that if the topic entity of a page does not appear in our KB, we cannot hope to have identified it at this point, but given the volume of text on a typical webpage, it is likely that there was a spurious match against some entity in the KB. Even if the topic is in our KB, it is possible that another entity was identified as the candidate topic if we had little information about the true topic in the KB. This motivates us to further apply a filtering step before assigning topics leveraging global formatting patterns.

\subsubsection{Global topic identification}

Our global topic identification step leverages the three observations we have made on semi-structured websites. 

\smallskip
\noindent
{\bf Step 1. Topic filtering by uniqueness:} According to our {\em Uniqueness} observation, if one entity is identified as the topic of many pages, it is unlikely to be the true topic. For example, if the word ``Help'' appears on every page and happens to match an entity in our KB, it may be identified as a candidate topic for many pages where the real topic does not exist in the KB. To counteract this problem, we discard any topic identified as a candidate topic of a large number of pages (\eg, $\geq 5$ pages).

\smallskip
\noindent
{\bf Step 2. Finding the dominant XPath:} Our {\em consistency} observation tells us that the text field containing the name or identifier of the topic entity should be in roughly the same location from page to page. We use this intuition to examine all candidate topics to try to deduce the areas of the page where they are most likely to occur.

For each page $w_{i}$, the XPaths to all mentions of topic candidates are collected  (note that the candidate topic may appear in multiple text fields on the page). For all candidate topics across the entire website $W$, we produce counts of how often each path appears. These paths are then ranked in descending order by count. If a path is highly ranked, there are many pages in which the candidate topic is found at that location on the page.

We then re-examine each page $w_{i}$ and find the highest-ranked path that exists on the page.
For all potential entities $j$ mentioned in that text field, the entity with the highest score $J(pageSet, entitySet_{j})$ as previously calculated is taken as the topic entity of the page. 

\smallskip
\noindent
{\bf Step 3. Webpage filtering by informativeness.} Finally, according to our {\em informativeness} observation, we only consider webpages with a predetermined minimum number of relation annotations (\eg, $\geq 3$); otherwise no topic is chosen for that page and it is discarded from the annotation process.  

\smallskip
\noindent
{\bf Parameter setting.} We note that at a few places we need to set parameters such as when to discard a topic if it is identified as the topic entity for many pages, and when to discard a webpage if the number of annotations is too small. Our goal is to filter obvious noise and expect our learning algorithm to be robust to remaining noise. So we set such parameters according to our empirical observations and tend to set a small value. 


\subsection{Relation annotation}
\label{sec:full-page-annotation}

\begin{algorithm}[t]
\caption{Annotating a full page for a predicate}\label{alg:full_page}
\label{alg:relation}
\begin{algorithmic}[1]
\small
\Require{A semi-structured webpage $w$, a set of predicate objects $O$, a set $C$ of clusters of XPaths for all predicate mentions across the website, and a list of predicates frequently duplicated.}
\Ensure{An annotation for each $o_i$ in $O$. }
\Procedure{bestLocalMention}{mentions}
\State $bestCount \gets 0$
\State $bestMentions \gets []$
\For{$mention \in mentions$}
\State {$ancestorNode$ $\gets$ highest level node containing} 
\State {\hspace{8ex}$mention$ and no other element in $mentions$}
\State {$neighborCount$ $\gets$ count of all objects for }
\State {\hspace{8ex}predicate under $ancestorNode$}
\If{$neighborCount > bestCount$}
\State $bestCount \gets neighborCount$
\State $bestMentions \gets [mention]$
\ElsIf{$neighborCount == bestCount$}
\State $bestMentions.append(mentions)$
\EndIf
\EndFor
\State \textbf{return} bestMentions
\EndProcedure
\\
\State {/* Main procedure for annotating $o_i$ in $O$. */}
\State $A \gets \emptyset$ /* initialize set of annotations */

\ForAll{$o_i$ in $O$}
\State {$mentions \gets $ XPaths corresponding to mentions}
\State {\hspace{8ex}of $o$ on $w$}
\State {$bestLocMen \gets bestLocalMention(mentions)$}
\If{count($bestLocMen$) == 1}
\State $mention \gets bestLocMen[0]$
\Else
\If{predicate is frequently duplicated}
\State {$mention \gets$ mention in $bestLocMen$}
\State {\hspace{8ex}corresponding to largest cluster in $C$}
\Else
\State $mention \gets Null$
\EndIf
\EndIf
\State $A.append(mention)$
\EndFor
\State \textbf{return} A
\end{algorithmic}
\end{algorithm}

Topic identification finds a topic for a set of webpages; at this point, we have retrieved from the KB all triples related to the topic entity and located all potential mentions of their objects on the page. The next step is to annotate those object mentions with the relations the webpage asserts.

If an object has only one mention on the page, and participates in only one triple with the topic entity, we can simply label that mention with the relation in the triple. However, that is often not the case. A traditional approach to distant supervision would have us annotate all object mentions with all predicates they represented; however, the DOM setting again presents challenges and opportunities that motivate a more thoughtful process. 

There are two potential problems. First, the object may participate in multiple relations with the topic, and all of these relationships may be represented on the page. If there is frequent overlap between two predicates (e.g., writers and directors of movies are often the same person), and they are mentioned in close proximity on the page, it will be hard to train a model to distinguish them. 

Second, the object may have relationships with entities other than the topic. Even though the majority of the information on a detail page is about the topic entity, there are often relevant entities shown at the side or bottom of the pages as recommendations. Such relevant entities often share common objects with the topic entity; for example, in Figure \ref{fig:DoTheRightThing}, the {\em Comedy} genre applies to both the topic movie {\em Do the Right Thing}, and to a recommended movie {\em Crooklyn}. However, other genres that apply to {\em Crooklyn} may not apply to the topic movie and blindly extracting them may cause mistakes. 

Since we emphasize precision over recall for annotation, we annotate {\em no more than one} mention of each object for a predicate. We try to harness local and global information to derive the correct annotation. In case the information for some predicates is duplicated on a page, we may miss labeling these true instances; however, this is acceptable since there is little benefit to extract the same triple twice. We compare our relation annotation approach with the traditional approach in our experiments.

\subsubsection{Local evidence for filtering}

We first seek local evidence to resolve the ambiguity when an object may have multiple mentions on the webpage or may participate in multiple relations. Our intuition is that when multiple objects exist for a predicate, webpages are likely to put all these objects together on the page, such as in a list of names.

Given the DOM node for a ($pred$, $obj$) pair, we find other objects for the same  predicate. We identify their closest ancestor nodes that do not contain another mention of the same object
(Line 5 of Algorithm \ref{alg:relation}),
and choose the ancestor whose subtree contains the highest count of objects we find for the predicate
(Line 21). 
In case of a tie, we resort to global evidence, as we describe shortly. 

\begin{example}
In Figure \ref{fig:DoTheRightThing}, Spike Lee's name appears in both the director and writer sections of the page. It is not visible in the figure, but Lee also appears in the cast list of the film as well. Let us assume our KB contains triples representing all these relationships, as well as the film's relationship with several other actors. When we attempt to annotate Lee's ``acted in'' relationship, we will choose the mention in the cast list because it is in the same section of the page as other objects of the ``acted in'' predicate.

\end{example}

\subsubsection{Global evidence via clustering}
Recall that in the topic identification step, we follow the \textit{consistency} observation and rank XPaths by how frequently they contain candidate topics, and prefer those with higher count. The \textit{consistency} observation applies to relations as well; however, since some multi-valued predicates are often represented in long lists and tend to occur in parts of the page with a higher variety of formatting than the page topic, the XPaths of relation candidates may be sparser than they are for topic candidates. For this reason, we instead cluster the XPaths of all potential object mentions of a predicate across pages, and prefer those XPaths that appear in larger clusters.

We use an agglomerative clustering approach, where in each iteration we find two nodes with the closest distance, and merge the clusters they belong to, until we reach the desired number of clusters. The distance function between two DOM nodes is defined as the Levenshtein distance~\cite{levenshtein1966binary} between their corresponding XPaths. The desired number of clusters is set to the maximum number of mentions of a single object in a webpage, such that all mentions of an object on a page can be placed into separate clusters.

This clustering step is only used in two cases: 1) the local evidence is insufficient to make an annotation decision; and 2) the same object appears as a value of a predicate in more than half of the annotated pages (recall the {\em informativeness} observation). It allows us to maintain high precision in our annotations, while still ensuring that we make annotations on the majority of pages. 

\begin{example}
In Figure \ref{fig:DoTheRightThing}, the genres for the film are mentioned twice, once correctly and once in a section describing a different movie in a recommendation; this occurs frequently on IMDb. Both sections contain two genres, so there is a tie from local evidence. After clustering all potential Genre mentions across all IMDb pages, we find that the mentions at the top of the page fall into a larger cluster, since all pages have the correct genres listed in a similar location. The mentions lower down in the page fall into a smaller cluster, since the genres of related films only sometimes have perfect overlap with the topic film. As a result, we annotate the mentions on the top of the page and discard those at the bottom.
\end{example}

\section{Training and Extraction}
\label{sec:learning}
We treat DOM extraction as a probabilistic multi-class classification problem where the input is a DOM node and the output is the probability of the node indicating a particular relation; in other words, the classes consist of all predicates in our ontology, along with an ``OTHER'' class. A caveat is that the DOM node that contains the topic entity is considered as expressing the ``name'' relation for the topic. The challenges we face in training such an ML model mainly come from the automatically generated training data, which can be oftentimes noisy and incomplete.

\subsection{Training Examples}
Our annotation process produces positive training labels, but does not explicitly produce negative labels. For each webpage that has received annotations, we select randomly $r$ unlabeled DOM nodes to use as negative examples (our ``OTHER'' class) for each positive example. Following convention in distantly supervised text extraction, we choose $r=3$. 

Since our annotations are incomplete and it is possible that some unlabeled nodes actually do express a relation, we thus shall not select the negative labels completely randomly. If we have labeled multiple positive examples of a predicate on a page, and they differ only in indices of their XPaths, we consider it likely that they belong to a list of values, such as the list of cast members for a film. When generating negative examples, we exclude other nodes that differ from these positives only at these indices, since they are likely to be part of the same list.

\subsection{Training}
\label{sec:model}
Presumably we can train any ML model, such as logistic regression, random forest and SVM. 
We experimented with several classifiers, but ultimately found the best results by modeling the probability of a relation given a node belonging to class $k$ as a multinomial logistic regression problem:
\begin{equation*}
Pr(Y = k\mid X) = \frac{e^{\beta_{k0} + \beta_k^T X}}{1 + \sum\limits_{i=1}^{M} e^{\beta_{i0} + \beta_i^T X}}
\end{equation*}


\noindent
where $X$ is a vector of features representing the node, $\beta_k$ is the set of weights associated with relation class $k \in \{1, \ldots, M\}$, and $\beta_{k0}$ is a class-specific intercept term. We harness the DOM structure to create two types of features to represent each node.

\smallskip
\noindent
\textbf{Structural features:}
We extract features relating to the attributes of neighboring nodes of the target node, as presented by Gulhane \textit{et al.} in the Vertex project \cite{Gulhane2011WebscaleIE}. Specifically, we examine the node itself, ancestors of the node,  
and siblings of those ancestors (up to a width of 5 on either side) for the following HTML element attributes:
\texttt{tag}, \texttt{class}, \texttt{ID}, \texttt{itemprop}, \texttt{itemtype}, and \texttt{property}. For each attribute of these nodes, we construct features as a 4-tuple consisting of \textit{(attribute name, attribute value, number of levels of ancestry from the target node, sibling number)}.

\smallskip
\noindent
\textbf{Node text features:}
In addition to the structural features, we generate features based on the texts of nearby nodes. We 
compile a list of strings that appear frequently on the website and check if any of these strings appears nearby the node being classified; 
if so, a feature is created consisting of the word and the path through the tree to the node containing the string.

\subsection{Extraction}
In extraction, we apply the logistic regression model we learned to all DOM nodes on each page of the website. When we are able to identify the ``name'' node on a page, we consider the rest of the extractions from this webpage as objects and use the text in the topic node as the subject for those extracted triples. Our probabilistic extractor outputs a confidence value for each extraction; varying the confidence threshold required to make an extraction allows a trade-off between precision and recall.

\section{Experimental Evaluation}
\label{sec:experiment}
Three experiments show the success of our techniques in several different verticals and settings. First, we show that our technique achieves state-of-the-art results on multiple verticals and is competitive with top systems trained on manually annotated data. Second, we show the advantages of our full annotation process over a baseline via experiments on IMDb, a complex website with many relationships. Third, to demonstrate the performance of our system on real-world data, we extracted over a million facts from dozens of long-tail websites in the Movie vertical.  

\subsection{Data sets}

\subsubsection{SWDE}
\label{sec:swde_dataset}

\begin{table}[t]
\caption{Four verticals of the SWDE dataset used in evaluation.} 
\label{table:SWDE_description}
\smallskip
\large
\centering
\resizebox{\linewidth}{!}{%
\begin{tabular}{lcrL{0.8\columnwidth}} 
\toprule
\textbf{Vertical} & \textbf{\#Sites} & \textbf{\#Pages} & \textbf{Attribute} \\ 
\midrule
Book & 10 & 20,000 & title, author, ISBN-13, publisher, publication\_date \\
Movie & 10 & 20,000 & title, director, genre, rating \\
NBA Player & 10 & 4,405 & name, height, team, weight \\
University & 10 & 16,705 & name, phone, website, type \\
\bottomrule
\end{tabular}}
\vspace{-1em}
\end{table}

To evaluate the performance of our approach across a range of verticals, we employed a subset of the Structured Web Data Extraction dataset (SWDE)~\cite{Hao2011FromOT} as our testbed. SWDE comprises ground truth annotations of 4--5 predicates for 8 verticals with 10 websites in each vertical and 200--2000 pages for each site. In our evaluation, we used four verticals, namely Movie, Book, NBA Player, and University, which all have named entities and have sufficient overlap between sites such that a KB constructed from one site should contain at least some of the same entities as the others. An overview of the dataset is provided in Table \ref{table:SWDE_description}. 

For the Movie vertical, we use a seed knowledge base derived from a download of the IMDb database that powers the IMDb website, with an ontology based on Freebase \cite{Bollacker2008FreebaseAC}. Table \ref{table:PoC_Graph} gives a summary of this KB, which contains 85 million triples. For the rest of the three verticals, we arbitrarily chose the first website in alphabetical order from each vertical (\url{abebooks.com} for Books, \url{espn.com} for NBA Player, and \url{collegeboard.com} for University), and used its ground truth to construct the seed KB. 
We randomly selected half of the pages of each website to use for annotation and training and used the other half for evaluation; the confidence threshold was set at 0.5.
\subsubsection{IMDb}
\label{imdb_dataset}
\begin{table}[t]
\caption{Common entity types and predicates in the KB used to distantly supervise experiments for the Movie vertical.}
\label{table:PoC_Graph}
\smallskip
\small
\centering
\begin{tabular}{lcc} 
\toprule
\textbf{Entity Type} & \textbf{\#Instances} & {\bf \#Predicates}\\
\midrule
Person & 7.67M & 15\\
Film & 0.43M & 19\\
TV Series & 0.12M & 9\\ 
TV Episode & 1.09M & 18\\
\bottomrule
\end{tabular}
\vspace{-1em}
\end{table}

To evaluate our approach on complex websites that provide information for many relations, we crawled IMDb\footnote{Express written consent must be obtained prior to crawling IMDb; see \url{http://www.imdb.com/conditions}} in May 2017, obtaining two sets of webpages: one consisting of 8,245 semi-structured pages about movies, and the other consisting of 1,600 semi-structured pages about people.

The seed KB is the same as used for the SWDE Movie vertical described in Section \ref{sec:swde_dataset}. A ground truth set was generated using the Vertex++ extractor described in Section \ref{sec:experimental_setup}, and manually spot-checked to ensure correctness. As with SWDE, half the pages were used for annotation and training and the other half for evaluation, with a 0.5 confidence threshold.

\subsubsection{CommonCrawl movie websites}
\label{section:CommonCrawlData}

The SWDE dataset validated our ability to obtain information from major websites well-aligned to our seed KB. However, to build a comprehensive KB, we need to obtain information also from smaller niche sites from all over the world. To evaluate our approach more broadly, we extracted knowledge from a range of movie websites in CommonCrawl\footnote{\small \url{http://commoncrawl.org/}}. 

The CommonCrawl corpus consists of monthly snapshots of pages from millions of websites~\cite{CommonCrawl} on the Web. We started with a few well-known sites, including {\em rottentomatoes.com}, {\em boxofficemojo.com}, and {\em themoviedb.org}. Based on a Wikipedia list of the largest global film industries by admissions, box office, and number of productions\footnote{\small \url{https://en.wikipedia.org/wiki/Film_industry}}, we then issued Google searches for terms corresponding to these countries, such as ``Nigerian film database'' and recorded resulting sites that had detail pages related to movies. We also issued a few additional searches related to specific genres we thought may not be well-represented in mainstream sites, including ``animated film database'' and ``documentary film database''. After compiling our list of sites, we then checked CommonCrawl\footnote{\small For each site, we scanned the CommonCrawl indices for all monthly scrapes prior to January 2018 and downloaded all pages for the site from the scrape with the largest number of unique webpages. Note that these scrapes do not necessarily obtain all pages present on a site, so the retrieved pages represent only a subset of the full site.} and kept all sites with more than one hundred pages available. Our final list contains a broad mix of movie sites, including sites based around national film industries, genres, film music, and screen size. Most are in English, but the set also includes sites in Czech, Danish, Icelandic, Italian, Indonesian, and Slovak.

In general, most websites contain a diversity in their pages reflected in their use of distinct templates for different entity types. For example, in a given site such as Rotten Tomatoes, movie pages often appear structurally similar to each other, whereas they differ significantly from celebrity pages. Recall from Section~\ref{subsec:defn} that we apply the clustering algorithm described in~\cite{Gulhane2011WebscaleIE} in an attempt to recover the inherent groups of templates. Our extraction approach is then run atop these individual clusters. 

In our experience, we observed that the clustering algorithm does not always succeed in identifying the different page types. For example, out of the 73,410 Rotten Tomatoes pages we obtained, 71,440 were placed into a single cluster, including almost all of the semi-structured pages as well as unstructured pages. Note that this is a challenge for our process, since our method will have to learn an extractor capable of working on multiple very different page types, including unstructured pages. 

We used the same knowledge base created for the SWDE Movie vertical as the seed KB. Since we do not have labels for CommonCrawl, we create the ground truth by sampling 100 extracted triples from each site and manually checking their correctness. A triple is considered to be correct if it expresses a fact asserted on the page from which it was extracted. We do not attempt to verify the real-world accuracy of the webpage's assertion, nor do we confirm which text fields on the page provided the extraction. Note however that the ground truth cannot tell us which triples we failed to extract from the website (\ie, recall).

\subsection{Experimental setup}
\label{sec:experimental_setup}
\noindent
{\bf Implementations: } We implemented our proposed technique, a distantly supervised extraction approach, together with three baselines.
\begin{enumerate}\tightlist
  \item {\sc Vertex++}: We implemented the Vertex wrapper learning algorithm~\cite{Gulhane2011WebscaleIE}, which uses manual annotations to learn extraction patterns, expressed by XPaths. We further improved the extraction quality by using a richer feature set. Training annotations were manually crafted by one of the co-authors to ensure correctness. Note that because of the access to precise training annotations, {\sc Vertex++} presumably should obtain better results than distantly supervised approaches.  
  \item {\sc \demabbrev-Baseline}: This baseline operates on the original Distant Supervision Assumption; that is, annotations are produced for all entity pairs on a page that are involved in a triple in the seed KB. The extraction model is the same as that described in Section \ref{sec:model}. However, since there is no concept of a page topic in this setting, our annotation must identify a pair of subject-object nodes for a relation; to produce features for the pair, we concatenate the features for each node. Note that at extraction time, because we do not know the topic of a page, we need to examine all possible \textit{pairs} of DOM nodes, which is computationally infeasible; thus, we identify potential entities on the page by string matching against the KB.
  \item {\sc \demabbrev-Topic}: This method applies 
Algorithm~\ref{alg:topic} 
for topic identification, but then annotates all mentions of an object with all applicable relations, bypassing the relation annotation process described in 
Algorithm~\ref{alg:full_page}.
  \item {\sc \demabbrev-Full}: This is the method proposed by the paper, applying 
Algorithm~\ref{alg:topic} and Algorithm~\ref{alg:full_page} 
for annotation, and techniques in Section~\ref{sec:learning} for extraction.
\end{enumerate}

We implemented our algorithms in Python. We used the logistic regression implementation provided by Scikit-learn using the LBFGS optimizer and L2 regularization with C=1 \cite{scikit-learn}, and used Scikit-learn's agglomerative clustering implementation for the clustering step. We set parameters exactly as the examples given in the texts. All experiments were run on an AWS m4.2xlarge instance with 32 GB of RAM memory.

\smallskip
\noindent
{\bf Evaluation:}
Our primary metrics for evaluation are precision, recall, and F1 score, the harmonic mean of precision and recall.
\[precision = \frac{tp}{tp + fp} \quad  \quad recall = \frac{tp}{tp + fn} \]

\noindent
where $tp$ is the number of true positive extractions, $fp$ is the number of false positives, and $fn$ is the number of false negatives.

\begin{table}[t!]
\renewcommand{\arraystretch}{1.3}
\caption{Comparing with state-of-the-art DOM extraction systems, {\sc \demabbrev-Full} obtains highest F-measure on two verticals. Bold indicates the best performance. (F-measures of prior work are directly taken from the papers.)}
\label{table:SWDE_main}
\medskip
\centering
\begin{minipage}{20pc}
\resizebox{\linewidth}{!}{%
\begin{tabular}{L{0.35\columnwidth}C{0.15\columnwidth}rrrr} 
\toprule
\textbf{System} & \textbf{Manual Labels}& \textbf{Movie} & \textbf{NBA Player} & \textbf{University}  & \textbf{Book} \\ 
\midrule
Hao \textit{et al.}~\cite{Hao2011FromOT} & yes & 0.79 & 0.82 & 0.83 & 0.86 \\
XTPath \cite{Cohen2015SemiSupervisedWW} & yes & 0.94 & \textbf{0.98} & 0.98 &\textbf{0.97}\\
BigGrams \cite{Mironczuk2017TheBT} & yes & 0.74 & 0.90 & 0.79 & 0.78\\
LODIE-Ideal \cite{Gentile2015EarlyST} & no & 0.86 & 0.9 & 0.96 & 0.85\\ 
LODIE-LOD \cite{Gentile2015EarlyST} & no  & 0.76 & 0.87\textsuperscript{\textit{ a}} & 0.91\footnote{\small F1 of distantly supervised systems is calculated based on predicates that were present in the ontology of the KB used for annotation. The KB for \demabbrev-Topic and \demabbrev-Full did not include {\sf Movie.MPAA-Rating} because lacking seed data. LODIE-LOD did not include {\sf University.Type}, {\sf NBAPlayer.Height}, and {\sf NBAPlayer.Weight}.}  & 0.78\\
RR+WADaR \cite{Ortona2015WADaRJW} & no & 0.73 & 0.80 & 0.79 & 0.70\\
RR+WADaR 2 \cite{Ortona2016JointRF} & no & 0.75 & 0.91 & 0.79 & 0.71\\
Bronzi \textit{et al.}~\cite{Bronzi2013ExtractionAI} & no & 0.93 & 0.89 & 0.97 & 0.91\\
\midrule
Vertex++ & yes & 0.90 & 0.97 & \textbf{1.00} & 0.94\\
\demabbrev-Baseline & no & NA\footnote{\small Could not complete run due to out-of-memory issue.} & 0.78 & 0.72 & 0.27\\
\demabbrev-Topic & no & \textbf{0.99}\textsuperscript{\textit{ a}} & 0.97 & 0.96 & 0.72\\
\demabbrev-Full & no & \textbf{0.99}\textsuperscript{\textit{ a}} & \textbf{0.98} & 0.94 &0.76\\
\bottomrule
\end{tabular}}
\end{minipage}
\vspace{-1.5em}
\end{table}

\subsection{Results on SWDE}
\label{sec:swde_results}
Table~\ref{table:SWDE_main} compares {\sc \demabbrev-Full} with our baselines, and the state-of-the-art results on {\em SWDE} in the literature. Among those systems, Hao {\em et al.}~\cite{Hao2011FromOT}, {\sc XTPath}~\cite{Cohen2015SemiSupervisedWW}, {\sc BigGrams}~\cite{Mironczuk2017TheBT}, and {\sc Vertex++} use manual annotations for training; {\sc LODIE-Ideal} and {\sc LODIE-LOD} ~\cite{Gentile2015EarlyST} conduct automatic annotation ({\sc LODIE-Ideal} compares between all web sources in a vertical and {\sc LODIE-LOD} compares with Wikipedia); {\sc RR+WADaR} (2)~\cite{Ortona2015WADaRJW} and Bronzi \textit{et al.}~\cite{Bronzi2013ExtractionAI} applied unsupervised learning. 

To compare with prior work, metrics in Table~\ref{table:SWDE_main} follows the methodology of Hao \textit{et al.}~\cite{Hao2011FromOT} in evaluating precision and recall on the SWDE dataset. Because their system can extract only one field per predicate per page, they base their metrics on \textit{page hits} for each predicate, giving credit for a page if at least one text field containing an object for that predicate is extracted. For fair comparison, we restrict our system to making one prediction per predicate per page by selecting the highest-probability extraction. We also present full results showing precision and recall across all mentions in Table~\ref{table:SWDE_detail}.

The {\sc \demabbrev-Full} automatic annotation process annotated a large percentage of pages when there was good overlap with the KB, producing at least one annotation on 75\% of pages for Movie, 97\% for NBAPlayer, 74\% for University, and 11\% for Book. {\sc \demabbrev-Full} achieved the best results on two of the four verticals, and overall obtained better results than most other system, even those using manual annotations. The other systems boasting vertical-best scores were supervised: XTPath allows up to 10\% of data to be used in training, while Vertex++ required two pages per site.

{\sc \demabbrev-Baseline} performed poorly across all verticals. In the movie vertical, where we used a large seed KB constructed from IMDb, {\sc DB-Baseline} produced too many annotations and ran out of memory, even with 32 GB of RAM. {\sc \demabbrev-Topic} performed similarly to {\sc \demabbrev-Full}. However, we note that many of the predicates in SWDE, such as {\sf University.Phone} and {\sf Book.ISBN-13}, have little ambiguity; as we will see soon in Section~\ref{sec:IMDB_results}, {\sc \demabbrev-Topic} performs worse in a more complex setting.

\begin{table}[t!]
\caption{Comparing with the supervised extractor {\sc Vertex++}, our distantly supervised system {\sc \demabbrev-Full} obtains comparable precision and recall across all extractions on the SWDE dataset. Bold instances indicate {\sc \demabbrev-FULL} beats {\sc Vertex++}.} 
\label{table:SWDE_detail}
\medskip
\centering
\resizebox{\linewidth}{!}{%
\begin{tabular}{llrrrrrr} 
\toprule
\multirow{2}{*}{\textbf{Vertical}} & \multirow{2}{*}{\textbf{Predicate}} &  \multicolumn{3}{c}{\textbf{Vertex++}} &  \multicolumn{3}{c}{\textbf{\demabbrev-Full}} \\
\cmidrule(lr){3-5}
\cmidrule(lr){6-8}
& & P & R & F1 & P & R & F1 \\
\midrule
\multirow{3}{*}{Movie}	& Title 	  & 1.00 & 1.00 & 1.00 & 1.00 & 1.00 & 1.00\\
						& Director 	  & 0.99 & 0.99 & 0.99 & 0.99 & 0.99 & 0.99\\
						& Genre 	  & 0.88 & 0.87 & 0.87 & \textbf{0.93} & \textbf{0.97} & \textbf{0.95}\\
						& MPAA Rating & 1.00 & 1.00 & 1.00 & NA & NA & NA\\
                        \cmidrule{2-8}
                        & Average & 0.97 & 0.97 & 0.97 & 0.97 & {\bf 0.99} & {\bf 0.98}\\
                        
\midrule  
\multirow{4}{*}{NBAPlayer} & Name 	& 0.99 & 0.99 & 0.99 & \textbf{1.00} & \textbf{1.00} & \textbf{1.00}\\
						   & Team 	& 1.00 & 1.00 & 1.00 & 0.91 & 1.00 & 0.95\\
						   & Weight & 1.00 & 1.00 & 1.00 & 1.00 & 1.00 & 1.00\\
						   & Height & 1.00 & 1.00 & 1.00 & 1.00 & 0.90 & 0.95\\
                           \cmidrule{2-8}
                           & Average & 1.00 & 1.00 & 1.00 & 0.98 & 0.98 & 0.98\\
\midrule
\multirow{4}{*}{University} & Name 	  & 1.00 & 1.00 & 1.00 & 1.00 & 1.00 & 1.00\\
							& Type 	  & 1.00 & 1.00 & 1.00 & 0.72 & 0.80 & 0.76\\
							& Phone	  & 0.97 & 0.92 & 0.94 & 0.85 & {\bf 0.95} & 0.90\\
							& Website & 1.00 & 1.00 & 1.00 & 0.90 & 1.00 & 0.95\\
                            \cmidrule{2-8}
                            & Average & 0.99 & 0.98 & 0.99 & 0.87 & 0.94 & 0.90\\
\midrule
\multirow{4}{*}{Book} & Title 			 & 0.99 & 0.99 & 0.99 & {\bf 1.00} & 0.90 & 0.95\\
					  & Author 			 & 0.97 & 0.96 & 0.96 & 0.72 & 0.88 & 0.79\\
					  & Publisher 		 & 0.85 & 0.85 & 0.85 & {\bf 0.97} & 0.77 & {\bf 0.86}\\
					  & Publication Date & 0.90 & 0.90 & 0.90 & {\bf 1.00} & 0.40 & 0.57\\
					  & ISBN-13 		 & 0.94 & 0.94 & 0.94 & {\bf 0.99} & 0.19 & 0.32\\
                      \cmidrule{2-8}
                      & Average & 0.93 & 0.93 & 0.93 & {\bf 0.94} & 0.63 & 0.70\\
\bottomrule
\end{tabular}
}
\vspace{-1em}
\end{table}

\begin{figure}[t!]
\centering
\includegraphics[scale=0.23]{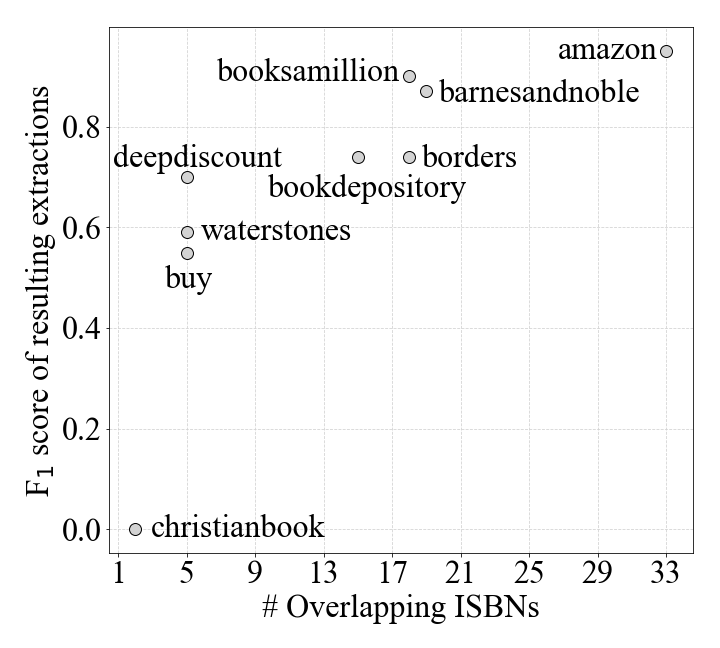}
\vspace{-1.5em}
\caption{Extraction F1 vs. \# of ISBNs overlapping w. the seed KB (potential number of annotated pages): lower overlap typically corresponds to lower recall. We omit acebooks.com site, which serves as the basis for the KB. An undefined F1 (when no extraction is produced) is shown as zero.}
\label{fig:swde_book_comparison}
\vspace{-1em}
\end{figure}

\begin{figure}[t!]
\centering
\includegraphics[scale=0.32]{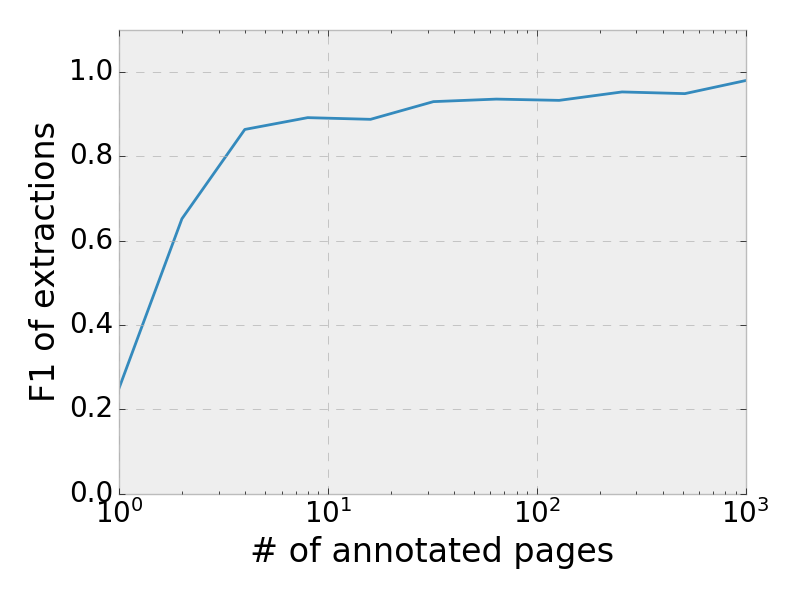}
\vspace{-1.5em}
\caption{Extraction F1 vs. \# of annotated pages used in the learning process for the SWDE Movie vertical. Note the log scale of the x-axis.}
\label{fig:swde_movie_annotation_count_plot}
\vspace{-1.2em}
\end{figure}

Table~\ref{table:SWDE_detail} compares {\sc \demabbrev-Full} with {\sc Vertex++} in detail on full extractions. We observe that {\sc \demabbrev-Full} obtains extraction quality comparable to {\sc Vertex++} most of the cases, and sometimes even higher precision. In particular, on predicates such as {\sf Movie.Genre} and {\sf University.Phone}, which frequently have two or more values, we achieve recall over 0.95, even higher than the state-of-the-art annotation-based supervised extractor.

{\sc \demabbrev-Full} performed best on the verticals where more annotations were produced, particularly the Movie and NBAPlayer verticals (note that we did not extract {\sf MPAA Rating} because our KB does not contain any triple with this predicate). In contrast, we obtained our lowest F-measure on the Book vertical. This is because in the Book vertical there is very little overlap between the seed KB, built from the ground truth for \url{acebooks.com}, and the other websites. As shown in Figure~\ref{fig:swde_book_comparison}, four of the sites had 5 or fewer pages representing books existing in our KB. However, even when we are able to annotate at most 5-20 webpages out of the 1000 pages, {\sc \demabbrev-Full} still obtained high precision on these sites. Figure \ref{fig:swde_movie_annotation_count_plot} shows that in the Movie vertical, a similar pattern emerges if a limitation is placed on the number of annotated pages that are used for learning the extractor model.

False positive extractions largely fell into two categories. First, the system sometimes extracted a node adjacent to the correct node, such as extracting the string ``Author:'' for the {\sf Book.Author} predicate. The annotations were correct, but the features of the nodes were too similar for the classifier to distinguish. One could imagine applying similar global filtering according to the {\em Informativeness} observation on extractions. Second, we still make annotation errors: in one website in the University vertical, all pages of the site listed the two potential {\sf University.Type} values (``public'' and ``private'') in a search box on every page. This produced incorrect annotations and bad resulting extractions. Ignoring certain areas of the pages like search box is likely to alleviate the problem.

\begin{table}[t!]
\caption{On IMDb, {\sc \demabbrev-Full} obtains much higher extraction quality than {\sc \demabbrev-Topic}.} 
\label{table:IMDB_experiments}
\medskip
\centering
\resizebox{\linewidth}{!}{%
\begin{tabular}{llrrrrrr} 
\toprule
\multirow{2}{*}{\textbf{Domain}} & \multirow{2}{*}{\textbf{Predicate}} &  \multicolumn{3}{c}{\textbf{\demabbrev-Topic}} &  \multicolumn{3}{c}{\textbf{\demabbrev-Full}} \\
\cmidrule(lr){3-5}
\cmidrule(lr){6-8}
& & P & R & F1 & P & R & F1 \\
\midrule
\multirow{7}{*}{Person} & name  & 1.0 & 1.0 & \textbf{1.0} & 1.0 & 1.0 & \textbf{1.0}\\
& alias  & 0.06 & 1.0 & 0.11 & 0.98 & 1.0 & \textbf{0.99}\\
& place of birth & 0.96 & 0.87 & 0.91 & 1.0 & 0.93 & \textbf{0.96}\\ 
& acted in  & 0.41 & 0.64 & 0.50 & 0.93 & 0.65 & \textbf{0.77}\\
& director of  & 0.48 & 0.92 & 0.63 & 0.95 & 0.95 & \textbf{0.95}\\
& writer of & 0.32 & 0.56 & 0.41 & 0.89 & 0.69 & \textbf{0.78}\\
& producer of& 0.48 & 0.24 & 0.32 & 0.8 & 0.44 & \textbf{0.57}\\
\cmidrule{2-8}
& All Extractions & 0.36 & 0.65 & 0.46 & 0.93 & 0.68 & 0.79\\
\midrule
\multirow{10}{*}{Film/TV} & title & 1.0 & 1.0 & \textbf{1.0} & 1.0 & 1.0 & \textbf{1.0}\\
& has cast member  & 0.93 & 0.46 & 0.62 & 1.0 & 0.49 & \textbf{0.66}\\
& directed by& 0.80 & 0.99 & 0.88 & 0.93 & 0.98 & \textbf{0.95}\\
& written by & 0.99 & 0.67 & 0.80 & 0.99 & 0.89 & \textbf{0.94}\\
& release date  & 0.37 & 0.14 & 0.20 & 1.0 & 0.63 & \textbf{0.77}\\
& release year  & 0.74 & 0.96 & 0.84 & 0.91 & 1.0 & \textbf{0.95}\\
& genre  & 0.80 & 1.0 & 0.89 & 1.0 & 0.99 & \textbf{0.99}\\
& (TV) episode number & 1.0 & 1.0 & \textbf{1.0} & 1.0 & 1.0 & \textbf{1.0}\\
& (TV episode) season number  & 0.98 & 1.0 & \textbf{0.99} & 0.87 & 1.0 & 0.93\\
& (TV episode) series  & 0.50 & 0.01 & 0.02 & 1.0 & 1.0 & \textbf{1.0}\\
\cmidrule{2-8}
& All Extractions & 0.88 & 0.59 & 0.70 & 0.99 & 0.65 & 0.78\\
\bottomrule
\end{tabular}
}
\vspace{-1em}
\end{table}

\begin{table}[t]
\caption{Accuracy of the automated annotations on IMDb Person and Film domains, with 20 pages scored per predicate. Recall is measured as the fraction of facts from KB that were correctly annotated. Comparing with {\sc \demabbrev-Topic}, {\sc \demabbrev-Full} has higher precision at the cost of slightly low recall.} 
\label{table:IMDB_annotation_results}
\medskip
\centering
\resizebox{\linewidth}{!}{%
\begin{tabular}{llrrrrrr} 
\toprule
\multirow{2}{*}{\textbf{Domain}} & \multirow{2}{*}{\textbf{Predicate}} &  \multicolumn{3}{c}{\textbf{\demabbrev-Topic}} &  \multicolumn{3}{c}{\textbf{\demabbrev-Full}} \\
\cmidrule(lr){3-5}
\cmidrule(lr){6-8}
& & P & R & F1 & P & R & F1 \\
\midrule
\multirow{7}{*}{Person} & alias  & 0.19 & 1.00 & 0.33 & 1.00  & 0.71 & 0.83\\
& place of birth & 0.84 & 0.55 & 0.67 & 0.90 & 0.45 & 0.60\\ 
& acted in  & 0.63 & 0.99 & 0.77 & 0.98 & 0.83 & 0.90\\
& director of  & 0.26 & 0.99 & 0.41 & 0.88 & 0.36 & 0.51\\
& writer of & 0.33 & 0.99 & 0.5 & 0.77 & 0.81 & 0.79\\
& producer of & 0.45 & 0.98 & 0.61 & 0.55 & 0.91 & 0.68\\
\cmidrule{2-8}
& All Annotations & 0.46 & 0.99 & 0.60 & 0.93 & 0.78 & 0.83\\
\midrule
\multirow{10}{*}{Film/TV} & has cast member  & 0.83 & 0.88 & 0.86 & 0.99 & 0.80 & 0.89\\
& directed by& 0.47 & 0.74 & 0.58 & 0.88 & 0.71 & 0.79\\
& written by & 0.68 & 0.52 & 0.59 & 0.90 & 0.36 & 0.51\\
& release date  & 0.53 & 0.59 & 0.56 & 1.0 & 0.56 & 0.72\\
& release year  & 0.27 & 0.75 & 0.39 & 1.0 & 0.71 & 0.83 \\
& genre  & 0.55 & 0.82 & 0.66 & 0.96 & 0.82 & 0.88\\
& (TV) episode number & 0.45 & 0.25 & 0.32 & 1.0 & 0.20 & 0.33\\
& (TV episode) season number  & 0.89 & 0.40 & 0.55 & 0.88 & 0.35 & 0.50\\
& (TV episode) series  & 0.44 & 0.42 & 0.43 & 1.0 & 0.42 & 0.59\\
\cmidrule{2-8}
& All Annotations & 0.53 & 0.80 & 0.61 & 0.96 & 0.71 & 0.83\\
\bottomrule
\end{tabular}
}
\vspace{-1.2em}
\end{table}

\begin{table}[t]
\caption{Accuracy of topic identification on the IMDb dataset.}
\label{table:topic_identification}
\medskip
\centering
\small
\begin{tabular}{lrrr} 
\toprule
\textbf{Domain} & \textbf{P} & \textbf{R} & \textbf{F1} \\
\midrule
Person & 0.99 & 0.76 & 0.86 \\
Film/TV & 0.97 & 0.88 & 0.92 \\
\bottomrule
\end{tabular}
\vspace{-1.5em}
\end{table}

\subsection{Results on IMDb}
\label{sec:IMDB_results}

The IMDb dataset is a challenging testbed that demonstrates why our annotation scheme is critical to obtaining high extraction quality when extracting for complex domains. In particular, we compare the extraction quality of {\sc \demabbrev-Full} with {\sc \demabbrev-Topic} in Table~\ref{table:IMDB_experiments}, and compare their annotation quality in Table~\ref{table:IMDB_annotation_results}. We note that our KB is constructed based on a download of IMDb data, so there is overlap between our KB and IMDb's website\footnote{Note, however, that the seed KB includes only a subset of facts found on the website (notably, it only contains links between people and movies if the person is a "principal" member of the film such as a lead actor or highly-billed writer or producer.) For example, for the pages in our web crawl, about 14\% of {\sf has cast member} facts on the webpages are present in the KB, along with 9\% of {\sf producer of}, 38\% of {\sf director of} and 58\% of {\sf genre}. This means the KB is biased toward certain types of entities. This can complicate extraction in the case that this bias correlates with presentation on the webpage; however, this approximates the case of a seed KB containing only ``popular'' entities in a setting where we want to extract all entities, including those in the long tail.}. The relationship of the KB and dataset also meant that we had strong keys connecting an IMDb page to a KB entity for most, but not all, of the pages, giving us ground truth to evaluate our topic identification step on this subset of pages. As shown in Table \ref{table:topic_identification}, we were very successful in identifying page topics.

One challenging aspect of IMDb is that many of the predicates, such as {\sf has cast member} and {\sf director of}, are multi-valued and consist of long lists of values (often 20 or more). Multi-valued predicates are a challenging and little-explored topic in semi-structured extraction, with most prior unsupervised and semi-supervised extractors restricting the problem to single-valued extractions \cite{Bronzi2013ExtractionAI,Hao2011FromOT,Gentile2015EarlyST,Gulhane2010ExploitingCR,Gulhane2011WebscaleIE}. In addition, IMDb contains many sections of the page that could easily produce false annotations (and thus extractions). For example, all {\sf People} pages include a ``Known For'' section listing the person's four most famous films; however, this section doesn't correspond to a particular predicate, so any system that learns to extract it will produce erroneous extractions. There are several similar sections that contain names of related entities but do not correspond to a predicate, such as the ``People who liked this also like'' list visible in Figure \ref{fig:DoTheRightThing}.

On this website, {\sc \demabbrev-Full} obtains very high precision (99\% on Film/TV pages, 93\% on Person pages) and reasonable recall (65\% and 68\%). The fairly low recall is mainly dominated by false negatives for {\sf acted in} and {\sf has cast member}, which contribute a large number of triples (non-weighted recall among all predicates is 81\% and 90\% instead). Lower recall on these predicates, as well as { \sf producer of} and { \sf writer of}, is partially due to the fact that these are multi-valued predicates. In this case, on the {\sf acted in} predicate, the recall loss is exacerbated by the fact that our seed KB only contains actors when associated IMDb character information is available, which is represented with a specific feature on the website. Lower precision on {\sf producer of} and { \sf writer of} is also due to the challenging nature of these predicates, whose objects are frequently only present in extraneous fields of the page such as a ``Projects in Development'' section, causing incorrect annotations. While our results are imperfect, the significant accuracy gain over {\sc \demabbrev-Topic} shows the advantages of our technique.

Annotation has decent precision (96\% on Film/TV, 93\% on Person), though sometimes lower recall (71\% and 78\%); this is because our annotation algorithms strive for high precision. Based on the slightly noisy annotations, we are able to train a robust extractor model, achieving an extraction F1 of 11\% higher than \demabbrev-Topic on Film/TV, and 72\% higher on Person. 

In contrast, {\sc \demabbrev-Topic} obtains much lower annotation precision than {\sc \demabbrev-Full}, though the recall is slightly higher. The very noisy annotations lead to low quality extractors: for 7 predicates, the extraction precision is below 50\%. In particular, {\sc \demabbrev-Topic} fails on Person pages, where there is significant ambiguity as films are mentioned in sections like ``Known For'' and ``On Amazon Video'' as well as the filmography. On the {\sf alias} predicate, {\sc \demabbrev-Topic} struggled since variations of a person's name show up in many places on the page, such as the episode names of talk shows on which they appeared or character names.

\begin{table*}[t]
\caption{{\demcaps} obtains an average of 83\% precision on long-tail multi-lingual movie websites from CommonCrawl when using a 0.5 confidence threshold.}
\label{table:commoncrawl_results}
\medskip
\centering
\resizebox{\linewidth}{!}{%
\begin{tabular}{llrrrrR{0.15\linewidth}R{0.15\linewidth}r} 
\toprule
\textbf{Website} & \textbf{Focus} & \textbf{\# of Pages} & \textbf{\# of Annotated Pages} & \textbf{\# of Annotations} & \textbf{Total Extractions} & \textbf{Ratio of Extracted to Annotated Pages} & \textbf{Ratio of Extraction to Annotation} & \textbf{Precision} \\ 
\midrule
themoviedb.org & General film information & 32,143 & 10,182 & 113,302 & 347,690 & 1.87 & 3.07 & 1.00 \\
blaxploitation.com & Blaxploitation films & 670 & 274 & 553 & 1,182 & 2.40 & 2.14 & 1.00 \\
danksefilm.com & Danish films & 2,100 & 403 & 1,712 & 13,146 & 2.98 & 7.68 & 0.98 \\
archiviodelcinemaitaliano.it & Italian films & 1,573 & 617 & 2,734 & 13,135 & 2.52 & 4.80 & 0.98 \\
filmitalia.org & Italian films & 2,847 & 909 & 4,247 & 10,074 & 1.96 & 2.37 & 0.96 \\
kmdb.or.kr & Korean films & 1,351 & 29 & 137 & 389 & 3.14 & 2.84 & 0.95 \\
britflicks.com & British films & 1,464 & 721 & 3,944 & 4,306 & 1.32 & 1.09 & 0.92 \\
rottentomatoes.com & Film reviews & 73,410 & 18,685 & 82,794 & 410,012 & 2.86 & 4.95 & 0.91 \\
moviecrow.com & Indian films & 569 & 84 & 271 & 912 & 2.70 & 3.37 & 0.91 \\
nfb.ca & Canadian films & 39,780 & 2,130 & 9,802 & 67,428 & 5.28 & 6.88 & 0.91 \\
kinobox.cz & Czech films & 37,988 & 2,940 & 16,820 & 60,337 & 5.58 & 3.59 & 0.90 \\
samdb.co.za & South African films & 1,424 & 10 & 25 & 281 & 6.60 & 11.24 & 0.88 \\
dianying.com & Chinese films & 15,789 & 1,333 & 3,998 & 48,302 & 8.22 & 12.08 & 0.84 \\
giantscreencinema.com & IMAX films & 370 & 50 & 333 & 856 & 3.00 & 2.57 & 0.83 \\
myanimelist.net & Animated films & 5,588 & 644 & 3,513 & 55,904 & 7.00 & 15.91 & 0.80 \\
hkmdb.com & Hong Kong films & 6,350 & 741 & 3,491 & 43,486 & 4.19 & 12.46 & 0.75 \\
bollywoodmdb.com & Bollywood films & 1,483 & 167 & 671 & 3,132 & 4.14 & 4.67 & 0.72 \\
soundtrackcollector.com & Movie soundtracks & 4,192 & 1,446 & 6,714 & 24,032 & 2.38 & 3.58 & 0.70 \\
spicyonion.com & Indian films & 5,898 & 752 & 2,375 & 7,439 & 2.89 & 3.13 & 0.70 \\
shortfilmcentral.com & Short films & 32,613 & 2,610 & 5,188 & 87,100 & 12.47 & 16.79 & 0.69 \\
filmindonesia.or.id & Indonesian films & 2,901 & 577 & 2,198 & 9,178 & 3.83 & 4.18 & 0.67 \\
the-numbers.com & Financial performance & 74,767 & 23,173 & 141,145 & 430,594 & 2.23 & 3.05 & 0.65 \\
sodasandpopcorn.com & Nigerian films & 3,401 & 190 & 423 & 3,862 & 13.16 & 9.13 & 0.62 \\
christianfilmdatabase.com & Christian films & 2,040 & 712 & 5,420 & 33,127 & 2.20 & 6.11 & 0.59 \\
jfdb.jp & Japanese films & 1,055 & 69 & 341 & 1,683 & 2.42 & 4.94 & 0.58 \\
kvikmyndavefurinn.is & Icelandic films & 235 & 49 & 185 & 868 & 3.96 & 4.69 & 0.57 \\
laborfilms.com & Labor movement films & 566 & 124 & 445 & 4,969 & 4.38 & 11.17 & 0.45 \\
africa-archive.com & African films & 1,300 & 300 & 892 & 2,150 & 2.31 & 2.41 & 0.42 \\
colonialfilm.org.uk & Colonial-era films & 1,911 & 48 & 212 & 1,605 & 20.00 & 7.57 & 0.29 \\
sfd.sfu.sk & Slovak films & 1,711 & 61 & 140 & 1,727 & 20.05 & 12.34 & 0.21 \\
bcdb.com & Animated films & 912 & 17 & 44 & 0 & 0.00 & 0.00 & NA \\
bmxmdb.com & BMX films & 924 & 1 & 1 & 0 & 0.00 & 0.00 & NA \\
boxofficemojo.com & Financial performance & 74,507 & 2 & 4 & 0 & 0.00 & 0.00 & NA \\
\midrule
Total & - & 433,832 & 70,050 & 414,074 & 1,688,913 & 3.22 & 4.08 & 0.83 \\
\bottomrule
\end{tabular}}
\vspace{-1.5em}
\end{table*}

\begin{figure}[t!]
\centering
\includegraphics[scale=0.4]
{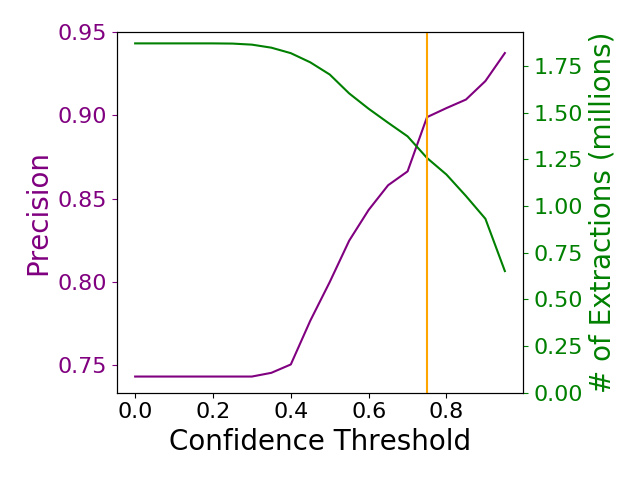}
\vspace{-1.5em}
\caption{Extraction precision vs. number of extractions on the CommonCrawl dataset at various confidence thresholds; the orange bar indicates a 0.75 threshold yielding 1.25 million extractions at 90\% precision.}
\label{fig:confidence_threshold}
\vspace{-1.2em}
\end{figure}

\begin{table}[t]
\caption{Number of annotations, extractions, and precision for the 10 most extracted predicates on the CommonCrawl dataset at a 0.5 confidence threshold.}
\label{table:predicate_counts}
\medskip
\centering
\resizebox{\linewidth}{!}{%
\begin{tabular}{lrrr} 
\toprule
\textbf{Predicate} & \textbf{\#Annotations} & \textbf{\#Extractions} & \textbf{Precision} \\
\midrule
film.hasCastMember.person & 78,527 & 441,368 & 0.98 \\
person.actedIn.film & 86,273 & 379,848 & 0.96 \\
film.hasGenre.genre & 40,359 & 175,092 & 0.90 \\
film.hasReleaseDate.date & 25,213 & 132,891 & 0.41 \\
film.wasDirectedBy.person & 25,159 & 85,244 & 0.94 \\
person.directorOf.film & 14,893 & 67,408 & 0.72 \\
person.createdMusicFor.film & 7,065 & 61,351 & 0.25 \\
person.hasAlias.name & 4,654 & 59,051 & 0.99 \\
film.wasWrittenBy.person & 18,643 & 58,645 & 0.93 \\
person.writerOf.film & 8,665 & 36,871 & 0.52 \\

\midrule
All Predicates & 414,194 & 1,688,913 & 0.83 \\
\bottomrule
\end{tabular}
}
\vspace{-1em}
\end{table}

\subsection{Results on CommonCrawl movie sites}
\label{sec:commoncrawl_results}
Figure \ref{fig:confidence_threshold} shows the precision and number of extractions from detail pages of all 33 movie websites at different thresholds for the extraction confidence; for example, a confidence threshold of 0.75 yields 1.25 million extractions at 90\% precision. Table~\ref{table:commoncrawl_results} shows a detailed breakdown of the 33 sites using an confidence threshold of 0.5; at this threshold, {\sc \demabbrev-Full} extracted nearly 1.7 million triples off a basis of 440,000 annotations, obtaining a 4:1 extraction to annotation ratio and 83\% precision. Table~\ref{table:commoncrawl_results} shows a detailed breakdown of the results on each site. 
We highlight several successes by our extraction on CommonCrawl.

\begin{itemize}\tightlist
  \item Our movie websites are carefully selected to include long-tail movie websites. The only exceptions are {\em themoviedb.org} and {\em rottentomatoes.com}, which provide general film information, and where we obtain extraction precisions of 100\% and 91\%, respectively. For the rest of the websites, we obtain a precision over 80\% on half of them. As shown in Table~\ref{table:predicate_counts}, for 6 out of the top-10 most extracted predicates, our overall precision is above 90\%. 
  \item {\sc \demabbrev-Full} obtains greater than 90\% precision on some of the foreign-language sites in Italian ({\em filmitalia.org}), Danish ({\em danksefilm.com}), and Czech ({\em kinobox.cz}).
  \item {\sc \demabbrev-Full} was able to obtain high precision on sites when we can annotate only on a few tens of webpages, such as {\em kmdb.or.kr} and {\em moviecrow.com}.
  \item Precision increases monotonically with respect to confidence, allowing for a trade-off between precision and recall.
  \item {\sc \demabbrev-Full} is able to extract triples that include entities not present in the seed KB. At 0.5 confidence, the ratio between annotated topic entities and extracted entities is 1:3.22, where most newly extracted entities are long tail entities.
  \item One notable instance of an inability to extract being a good thing occurred on boxofficemojo.com. Despite contributing over 74,000 pages to our dataset, the pages in the CommonCrawl scrape did not include any detail pages about movies, instead consisting almost entirely of daily box office charts. We consider the fact that our system did not produce erroneous extractions from this site to be a success.
\end{itemize}

\noindent
{\bf Comparison with Knowledge Vault~\cite{Dong2014KnowledgeVA}:} Knowledge Vault trained two DOM extractors to extract knowledge from the Web using Freebase as the seed KB, and the precision when applying a threshold of 0.7 is 0.63 and 0.64 respectively~\cite{Dong2014FromDF}. Because there are no broken-out details provided about the performance, we cannot conduct a side-by-side fair comparison. However, we believe {\demcaps} obtained significantly higher precision because (1) we obtained an average precision of 0.94 on mainstream websites in SWDE, even with quite limited training data on some of the domains, and (2) on long-tail multi-lingual websites in CommonCrawl, which present significant challenges for extraction, we obtained an average precision of 0.83. In addition, unlike Knowledge Vault, we allow extracting facts where the subjects and objects are not present in the seed database. On the limited set of CommonCrawl websites, we already extracted knowledge for 155K new entities.
  
\subsubsection{Discussion}
\label{sec:CommonCrawlDiscussion}
On the other hand, we also observe low accuracy in extracting some of the long-tail websites. We next discuss where our algorithm may fall short and potential improvement directions.

\smallskip
\noindent
{\bf Semantic ambiguity:} About a third of all mistakes happen in presence of semantic ambiguity for predicates. For example, \url{spicyonion.com} and \url{filmindonesia.or.id} list all films that a person is involved in, without distinguishing the role of the person like {\sf writer}, {\sf director}, or {\sf actor}; \url{the-numbers.com} contains long lists of the date and box office receipts for every day the film was in theaters, instead of just {\sf release dates}; \url{christianfilmdatabase.com} and \url{laborfilms.com} contain a list of all genres on every page, rather than just genres for the topic movie. Ideally in such cases we shall make no annotation, but because the area of the webpages contains a superset of values for a multi-valued predicate, our method will wrongly annotate and learn to extract that area. 

\smallskip
\noindent
{\bf Over-represented or under-represented types and predicates:} Most of the websites contain information solely about movies, but our KB contains over a million TV episodes (vs. 430K movies). Topic identification may wrongly match movies to episodes, such as in websites like \url{dianying.com}, \url{colonialfilm.org.uk}, \url{myanimelist}, and \url{samdb.co.za}, accounting for 5\% of total errors. On the other hand, for predicates such as {\sf sound editor} and {\sf camera operator}, which do not exist in the seed KB, we do not necessarily make annotation mistakes but can make extraction mistakes when the XPaths are very similar. This happened on \url{kvikmyndavefurinn.is}, \url{jfdb.jp}, and \url{sfd.sfu.sk}.

\smallskip
\noindent
{\bf Template variety:} Some websites, such as \url{colonialfilm.org.uk} and \url{bollywoodmdb.com}, may change the order of the predicates on the webpage, with the predicate indicated by an adjacent string. {\sc \demabbrev-Full} may fail to learn the correct pattern when the text features are not strong enough; this category accounted for 23\% of errors. We leave for future work to investigate how many of these aforementioned mistakes can be solved by applying knowledge fusion~\cite{Dong2014KnowledgeVA, Dong2014FromDF} on the extraction results.

\smallskip
\noindent
{\bf Disjoint webpages or non-detail pages:} Many websites contain pages generated by different templates, or even non-detail pages. Ideally, our clustering algorithm shall be able to separate them. Unfortunately, a strict implementation of Vertex clustering algorithm~\cite{Gulhane2011WebscaleIE} sometimes does not obtain ideal results, e.g., for \url{sodas\\andpopcorn.com}, so 36\% of bad extractions occurred on these pages. This causes confusion at the extraction time. Indeed, we manually removed non-detail pages in our evaluations; including them would increase the extraction count by 200,000, but reduce the precision to 0.74. A robust clustering algorithm is critical to apply our algorithm to the whole web.

\eat{
\begin{figure}[t!]
\includegraphics[width=8cm]{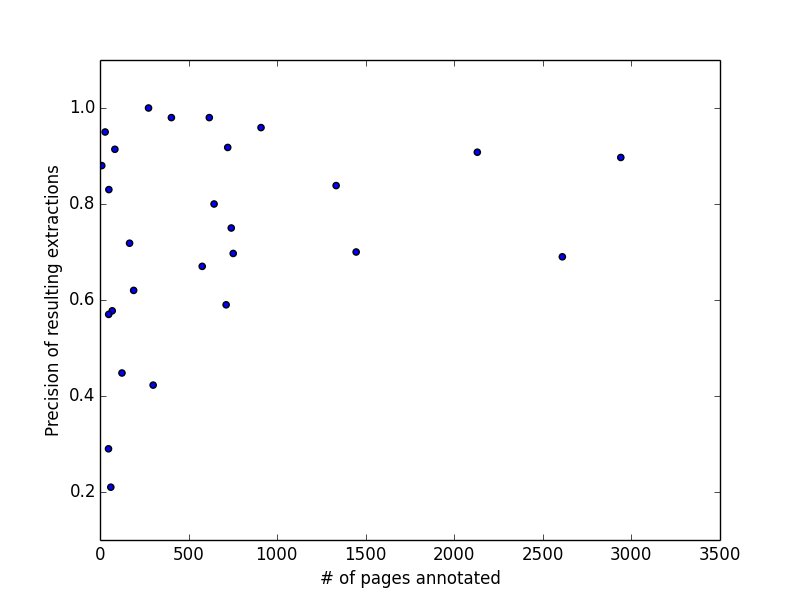}
\caption{The number of pages annotated vs. the precision of extractions for each site with fewer than 3500 annotated pages from the CommonCrawl dataset.}
\label{fig:commoncrawl_annotation_precision_graph}
\end{figure}
}

\section{Related Work}
\label{sec:related}

Relation extraction has been an active area of research in the Natural Language Processing community for decades. A recent overview of this area can be found in a survey by Grishman~\cite{grishman2012information}. Here we focus on techniques applicable to the semi-structured Web:

\smallskip
\noindent\textbf{Wrapper induction:} Early work for extracting semi-structured data was based on a supervised learning scheme called wrapper induction \cite{Kushmerick1997WrapperIF, zhai2007extracting, Gulhane2011WebscaleIE, muslea1999hierarchical}. The goal of wrapper induction is to learn wrappers (or rules) for relations of interest based on a sample of annotated data for a site. They work under the assumption that the observed HTML-formatted data is an instance of a database-backed template or schema, and the goal is to partially uncover these templates. Wrappers are often able to produce high quality data, i.e., extractions with high precision and often high recall. However they suffer from three problems: 1) they work only at the site-level and require human annotation effort for each site, 2) most wrappers were not designed to handle noisy annotations, so they tend to over-generalize in the presence of noise in the training data, and 3) they are often brittle --- they tend to break even due to minor changes in a site's structure.

Subsequent approaches aimed to alleviate some of these shortcomings. For example, in order to reduce annotation effort, Zhai \textit{et al.}~\cite{zhai2007extracting} used an active learning strategy to seek annotations from human editors when existing wrappers fail to extract. Dalvi \textit{et al.}~\cite{Dalvi2011AutomaticWF} proposed a generic framework to enable existing wrapper techniques be resilient to noisy annotations. Their method searches the space of wrappers induced by subsets of (noisy) annotations and aims to identify one that ranks high per wrapper quality that takes repeating patterns and schema size into account. 

Moreover, most wrappers were designed to work at the site level (except for \cite{wong2010learning}) which requires manual labor to create annotations for each site to be processed. Thus, a recent approach by Hao \textit{et al.}~\cite{Hao2011FromOT} promises to extract data for all sites in a vertical by only requiring that annotations be made for all pages of one site in that vertical. This approach entails accurate labeling for all pages of a site, for example by manually crafting regular expressions to extract attribute values of interest. Such demands of high-quality, site-wide labeling may not always be fulfilled and thus can be a severe limitation. In general, although wrappers result in high quality extractions, they are not scalable due to the annotation requirements.

\smallskip
\noindent\textbf{Automatic extraction:} Recent research activity has focused on devising unsupervised approaches for extraction, also known as \textit{automatic extraction}. The underlying assumption is that the HTML pages are a result of a small number of database-backed templates, and thus mining patterns in the DOM tree structure of web pages, the content and contexts should help uncover the templates. Approaches in this category \cite{arasu2003extracting, crescenzi2001roadrunner, zhai2005web, Gentile2015EarlyST} aim to either discover templates from a set of pages or find recurring patterns from a page to extract unseen relation instances from other pages. Although these approaches do not require any supervision, their major downside is that, typically, their extracted content requires an additional non-trivial post-processing step such as identification of attribute values, clustering, or ontology alignment.

\smallskip
\noindent\textbf{Seed KB based extraction:} Approaches in this category align attribute values in KB to text values on webpages to create annotations for learning wrappers. Gulhane \textit{et al.}~\cite{Gulhane2010ExploitingCR} exploit the redundancy in the attribute value content across websites in a vertical to make extractions from a new site in the vertical. They follow an iterative approach in which, at each iteration, some pages of a new site are annotated based on matching attribute values with the KB, and these annotated pages are used to learn wrappers for extracting from rest of the pages in the site. The LODIE project~\cite{Gentile2015EarlyST} is another example where annotations are generated based on matching attribute values on a page to dictionaries of attribute values pre-assembled from across the Linked Open Data cloud. Owing to their use of wrappers, both approaches require accurate annotations which can be unrealistic to expect for complex sites commonly found on the web.

An alternate approach is that used by DIADEM \cite{FGG+2014}, which, rather than assuming a KB, identifies extractable fields using a well-defined ontology and a set of  recognizers corresponding to each entity type as well as predicate labels. Unlike {\demcaps}, this requires manual work to define these recognizers for each predicate and for each language.

\smallskip
\noindent\textbf{Distant supervision based extraction:} More recently, research methods (including this work) have employed the distant supervision paradigm~\cite{Mintz2009DistantSF} for leveraging information in a KB as a source of supervision for creating potentially noisy annotations. However, these methods are mainly designed for unstructured data. Knowledge Vault (KV) project~\cite{Dong2014KnowledgeVA} is an exception however, which takes advantage of the DOM tree structure to predict relations for co-occurring entities on a page. KV requires the input pair of entities to exist in the KB, and hence cannot discover new entities. \demcaps, on the other hand, is able to discover new entities which is immensely useful for improving a KB's coverage.

Finally, for unstructured data, the strong distant supervision assumption has been relaxed to accommodate cases when a pair of entities may not hold for any relation in the KB~\cite{Riedel2010ModelingRA}, or may have multiple overlapping relations~\cite{Hoffmann2011KnowledgeBasedWS}. Most recently, {\em data programming}~\cite{ratner2016data} and {\em heterogeneous supervision}~\cite{liu2017heterogeneous} are proposed to unite diverse, possibly conflicting sources of supervision such as annotations by humans, supervision from a KB, or any labeling function.
These annotation heuristics could be applied to our approach.

\section{Conclusions}
\label{sec:conclude}

We have presented a new method for distantly supervised relation extraction from semi-structured websites. By using an entity-linking step in our annotation process to identify detail page topic entities, followed by a clustering process to identify the areas of pages corresponding to each predicate, we produce highly accurate annotations that can be used to train a supervised extractor. Experiments on the SWDE dataset demonstrate the state-of-the-art results on our system in multiple verticals, and a large-scale extraction project on hundreds of thousands of webpages shows the real-world usefulness of our approach. 

While we believe this work offers an important step in the direction of truly automatic extraction from the web, additional work is necessary to scale the process to extract from the whole web across multiple domains. This will involve methods to effectively identify semi-structured pages and correctly group them into template-based clusters, expansion of the topic identification process to deal with non-named-entities, and methods to automatically identify the domain of a website.

\balance


\bibliographystyle{abbrv}
\bibliography{vldb_sample}  










\end{document}